\documentclass[runningheads]{llncs}

\usepackage{graphicx}
\usepackage{tabularx,booktabs}
\usepackage[linesnumbered,ruled,longend]{algorithm2e}

\usepackage[T1]{fontenc}
\usepackage[utf8]{inputenc}
\usepackage[vietnamese,english]{babel}
\usepackage{lmodern}  
\usepackage{vntex}  

\usepackage{algpseudocode}
\usepackage{amsmath}
\usepackage{mathtools}
\usepackage{amssymb}
\usepackage{xcolor}
\usepackage{multirow}
\usepackage{comment}
\excludecomment{comment} 
\usepackage{array}
\usepackage{hyperref}
\usepackage{caption,graphicx,newfloat}
\newcolumntype{M}[1]{>{\centering\arraybackslash}m{#1}}
\begin{document}
\title{BERT-based model for Vietnamese Fact Verification Dataset
\thanks{
Bao Tran, T. N. Khanh, Khang Nguyen Tuong, Thien Dang and Quang Nguyen
contributed equally to this paper. \\ Corresponding author: \email{vthung@hcmut.edu.vn}
}}
%
%
\author{
Bao Tran\inst{1,2} 
\and T. N. Khanh \inst{1,2} \orcidID{0009-0007-2452-959X}
\and Khang Nguyen Tuong\inst{1,2}
\and Thien Dang\inst{1,2}
\and Quang Nguyen\inst{1,2}
\and Nguyen T. Thinh\inst{1, 2}
\and Vo T. Hung\inst{1,2}\orcidID{0000-0002-2910-1548}
}
\authorrunning{Bao et al.}
%
\institute{Faculty of Computer Science and Engineering, Ho Chi Minh City University of Technology (HCMUT), Vietnam \\
\email{\{bao.tran2003,vthung\}@hcmut.edu.vn}
\and Vietnam National University Ho Chi Minh City, Vietnam
}
%
\maketitle              
\begin{abstract}

The rapid advancement of information and communication technology has facilitated easier access to information. However, this progress has also necessitated more stringent verification measures to ensure the accuracy of information, particularly within the context of Vietnam.
This paper introduces an approach to address the challenges of Fact Verification using the Vietnamese dataset by integrating both sentence selection and classification modules into a unified network architecture. The proposed approach leverages the power of large language models by utilizing pre-trained PhoBERT and XLM-RoBERTa as the backbone of the network.
The proposed model was trained on a Vietnamese dataset, named ISE-DSC01, and demonstrated superior performance compared to the baseline model across all three metrics. Notably, we achieved a Strict Accuracy level of 75.11\%, indicating a remarkable 28.83\% improvement over the baseline model.


\end{abstract}

\section{Introduction}\label{sec:intro}
In the current era, the exponential growth of social media and online news platforms has led to an overwhelming increase in the volume of information available. 
Alongside this growth, the dissemination of misinformation has become widespread on the Internet, blurring the line between factual information and falsehoods. 
Consequently, the manual verification of a vast amount of information is impractical, as it requires significant human resources.
Hence, there is an urgent need for an accurate and automated mechanism for verification and fact-checking. 
This paper aims to address this need by focusing on enhancing the performance of fact-checking using a Vietnamese dataset.
Future research may explore the adaptation of this approach to other datasets and languages.
Natural Language Inference (NLI) is a field of study that examines the ability to draw conclusions about a hypothesis within the given context of a premise. 
Essentially, it aims to determine the logical relationship between a pair of text sequences.
These relationships can be classified into three main types: entailment, contradiction, and neutral. 
In an entailment scenario, the hypothesis aligns with the truth and can be inferred from the premise. 
On the other hand, contradiction arises when the negation of the hypothesis can be inferred from the premise. 
Lastly, in a neutral relationship, the logical connection between the hypothesis and the premise remains undetermined or ambiguous.

The benchmarks for Natural Language Inference (NLI) are well-represented by the Stanford Natural Language Inference (SNLI) \cite{DBLP:journals/corr/BowmanAPM15},  Multi-Genre Natural Language Inference (MultiNLI) \cite{DBLP:journals/corr/WilliamsNB17}, among others. 
In the domain of fact verification, several datasets have been developed to cater to the specific needs of this field. 
Notable examples include
FEVER \cite{DBLP:journals/corr/abs-1803-05355}, HOVER \cite{DBLP:journals/corr/abs-2011-03088}, SciFact \cite{DBLP:journals/corr/abs-2004-14974}, DanFEVER \cite{norregaard-derczynski-2021-danfever}. 
These datasets have played a crucial role in advancing research and development in the area of fact verification.



\textbf{Document Retrieval:}
The FEVER baseline employs the document retrieval module sourced from the DrQA system \cite{DBLP:journals/corr/ChenFWB17}. 
This module retrieves the top $k$ closest documents for a given query by calculating cosine similarity using binned unigram and bigram Term Frequency-Inverse Document Frequency (TF-IDF) vectors. 
Additionally, Rana {\textit{et al.}} (2022) \cite{DBLP:journals/corr/abs-2202-02646} proposed an improvement to this method by introducing a reduced abstract representation approach. 
This approach computes the TF-IDF similarity scores for all abstracts and focuses on the top-K similar abstracts.
On the other hand, Pradeep {\textit{et al.}} (2020) \cite{DBLP:journals/corr/abs-2010-11930} employed the BM25 scoring function, based on the Anserini IR toolkit \cite{DBLP:journals/corr/Wang17j}, to rank the abstracts from the corpus.

\textbf{Sentence Selection:}
In the FEVER approach, a basic sentence selection method is employed to organize sentences based on TF-IDF similarity to the claim. 
This method initially sorts the most similar sentences and then adjusts a threshold using validation accuracy on the development set.
An assessment is conducted on both DrQA and a straightforward unigram TF-IDF implementation to rank the sentences for selection.
Additionally, point-wise is also a simple and accessible approach. 
BEVER \cite{dehaven-scott-2023-bevers} employs a straightforward point-wise method for selecting sentences to generate the predicted evidence. 
The analysis considers two scenarios, treating the task as both a binary classification task and a ternary classification task.
The author of BERT \cite{DBLP:journals/corr/abs-1910-02655}, in addition to the point-wise method, also introduced the pair-wise method. 
This approach involves positive and negative sampling, followed by the application of rank scores to assess each instance.
Hinge Loss/Ranknet Loss is employed as the training criterion for this approach.

\textbf{Claim Verification:}
The FEVER document discusses the comparison of two models designed for recognizing textual entailment. 
In selecting a straightforward yet effective baseline, the authors opted for the submission by Riedel {\textit{et al.}} (2017) \cite{DBLP:journals/corr/RiedelASR17} from the 2017 Fake News Challenge. 
This baseline model is a multi-layer perceptron (MLP) featuring a single hidden layer, utilizing term frequencies and TF-IDF cosine similarity between the claim and evidence as key features.
Furthermore, the evaluation of the state-of-the-art in Recognizing Textual Entailment (RTE) involves the application of a decomposable attention (DA) model, specifically designed to establish attentional relationships between the claim and the evidence passage. 

Furthermore, there are various models employing different baselines, exemplified by models such as KGAT \cite{DBLP:journals/corr/abs-1910-09796}.
Notably, KGAT is designed akin to an undirected graph, where nodes gauge the significance of evidence in the text, and edges convey evidence to nodes, thereby enhancing the efficiency of verification.
Another example is the ProoFVer model \cite{DBLP:journals/corr/abs-2108-11357}, which utilizes a seq2seq model to generate inferences based on logical operations. 
These inferences encompass vocabulary variations between the assertion set and recursively derived evidence. 
Each inference is marked by a logical operator and determined based on the sequence of these operators.

However, research on fact verification for low-resource languages such as Vietnamese remains limited. 
To our knowledge, apart from the approach proposed by Duong {\textit{et al}} (2022) \cite{10013889}, which combined Knowledge Graph and BERT \cite{DBLP:journals/corr/abs-1810-04805} to verify facts on a private Vietnamese Wikipedia dataset, there has been no other published method for fact verification on Vietnamese dataset. 
There is a significant need to perform further research for fact verification on Vietnamese as the lack of prior work highlights the current knowledge gap. To fill this gap, we propose a different approach to perform fact verification on a public Vietnamese News dataset, from UIT Data Science Challenge\footnote{\label{DSC}\url{https://dsc.uit.edu.vn/}}.

Our contribution lies in the design of a pipeline capable of operating on the Vietnamese Fact Verification dataset and conducting a comparative analysis against a baseline.
Our approach integrates sentence selection and classification modules within a unified network architecture.
To address the limitations posed by the dataset's small size, our proposed approach harnesses the capabilities of large language models, specifically pre-trained PhoBERT and XLM-RoBERTa, as the backbone of the network.
The proposed model underwent training on a Vietnamese dataset known as ISE-DSC01 and exhibited superior performance in comparison to the baseline model across all three metrics.

The rest of the paper is as follows.
The section \ref{sec:methods} shows the approach.
The experimental setup and results are presented in section \ref{sec:experimental-results}.
Finally, section \ref{sec:conclusion} is the conclusion and discussion about future work.

\section{ISE-DSC01: A Vietnamese dataset for Fact Verification}\label{sec:dataset}

The dataset that we used for this study is ISE-DSC01 from UIT Data Science Challenge\footnotemark[3] contest from University of Information Technology - VNUHCM\footnote{\url{https://www.uit.edu.vn/}}.
The dataset was written in Vietnamese. 
According to the author of this dataset, the origin of the dataset is taken from some news websites in Vietnam. 

Given claim and evidence, the task requires to classify that claim as SUPPORTED, REFUTED, or NEI (Not Enough Information). 
If a claim is SUPPORTED or REFUTED, the system also needs to return a single evidence in corpus to convince the claim that supporting or refuting, otherwise the system doesn't need to return any evidence. 
This dataset is just like the FEVER\cite{DBLP:journals/corr/abs-1803-05355} dataset, but it has minor differences. 
About the FEVER\cite{DBLP:journals/corr/abs-1803-05355} dataset, before performing the evidence retrieval task, document retrieval task is needed to retrieve approriate document. 
Moreover, the FEVER dataset required to return the top 5 evidences that is nearly relevant to claim, instead of one. An example of a dataset with 4 items: Claim, Corpus, Evidence, and Label is also provided. 
Noted that, the evidence sentence demonstrate as a bold part of corpus. 


\begin{table}
\caption{Example dataset} \label{table:ex-dataset}
\tiny
\noindent \fbox{%
    \parbox{\textwidth}{%
        \textbf{Claim:} "Hàng trăm đơn đăng ký được hỗ trợ chi phí tàu xe của người lao động đã gửi đến Vietnam Airlines"\\
        \textit{Hundreds of applications that support for worker's transportation costs have been sent to Vietnam Airlines.}
        \\
        \textbf{Corpus} "...Sau khi chở lao động về Hà Nội tối qua, Vietnam Airlines hỗ trợ chi phí tàu xe cho người lao động về quê nhà. \textbf{Đại diện hãng cho biết đã tiếp nhận hàng trăm đơn đăng ký của người lao động}. Trong đó, có nhiều hoàn cảnh đặc biệt khó khăn như có người thân bị mắc bệnh hiểm nghèo, có người đã 7 năm rồi chưa được về quê..."\
        \textit{...After carrying workers to Hanoi last night, Vietnam Airlines supported transportation costs for workers to return their home. \textbf{The company said they had received hundreds of applications from workers}. Among them, there are many especially difficult situations such as suffering from a serious illness, some people have not been able to return home for 7 years...}
        \\
        \textbf{ID:} 20404
        \\
        \textbf{Label:} SUPPORTED
    }%
    
}
\end{table}

The ISE-DSC01 dataset contains a total of 49,675 news, split as a 38,684 train set, a 4,793 dev set, and a 5,396 test set. In train set, there are total 12,786 SUPPORTED label, 12,598 REFUTED, 13,309 NEI, which is almost balance.

\begin{figure}[!tb]
    \centering
    \includegraphics[width=0.85\linewidth]{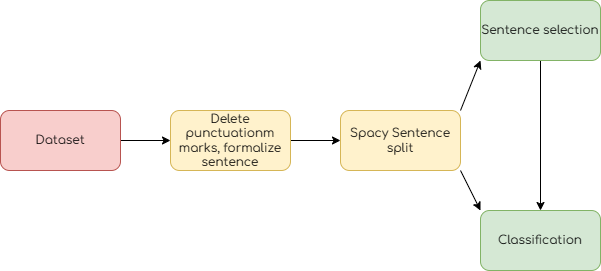}
    \caption{Data pipeline} \label{figure:data_pipeline}
    \label{fig:rightrecog}
\end{figure}

Figure \ref{figure:data_pipeline} shows the step of processing from the original dataset. 
Since the dataset is in the form of a document, but the input of the model is [$c$ [SEP] $s_i$], which will be mentioned in section \ref{sec:methods}, we need to split the corpus into a list of sentences.
Splitting a document into sentences is not an easy task because it has some special rules (e.g. After an ellipsis, if the first word is capitalized, we need to end the line, if not, we don't have to end the line).
By using the Spacy\footnote{https://spacy.io/} toolkit, we can split them into multiple sentences for almost every case, except for some really special cases. 

Furthermore, some punctuation that doesn't have any meaning is removed, and also convert capital letters into lowercase letters to formalize the document.
For rationale selection module, we use claim, evidence from the original dataset and the ground truth label. 
The label is 1 if the verdict of the dataset is \textit{SUPPORTED} of \textit{REFUTED}, otherwise the label is 0. 
With label classification module, the dataset has a quite different approach. 
With the SUPPORTED and REFUTED label, we utilize the claim and evidence from the orginial train dataset. 
And for the NEI label, to enhance the efficiency, we try to pick up the top-2 most relevant sentences with the claim in the corpus to create a dataset. 
The reason for this choice instead of top-1 selection is just to make the trainset balance out.

\section{Approach}\label{sec:methods}

In this section, we will describe our developed system for fact verification task.
Since the dataset is in Vietnamese, we need to further dissect the first sentence retrieval into two steps as the figure \ref{fig:rightrecog}.
Firstly, we want to retrieve the most relevant sentence in the corpus as shown in the figure \ref{fig:rightrecog}. 
For the label classification module, the sentence is classified as $\{SUPPORTED, REFUTED, NEI\}$ against the claim to give the final verdict. 

\subsection{Encoder}

We use BERT\cite{DBLP:journals/corr/abs-1810-04805}\footnote{We use BERT\cite{DBLP:journals/corr/abs-1810-04805} here for shortness, for each module, a different BERT version is used (e.g XLM-R\cite{conneau-etal-2020-unsupervised}, PhoBERT\cite{nguyen-tuan-nguyen-2020-phobert}, etc)} encoder to obtain the embeddings for each pair of claim sentence (denote as $c$) and each sentence in the corpus (denote as $s_i$). The input is described as:
\begin{center}
[CLS] $c_1$ $c_2$ ... $c_n$ [SEP] $w_{i1}$ $w_{i2}$ ... $ w_{ik}$ [SEP]
\end{center}
where $c_1,...c_n$ are word in claim sentence and $w_{i1},  ... ,w_{ik} \in s_i$ are word in cadidate sentence $s_i$. 
Here, a [SEP] token is added between two sentences to separate them. 
And a [CLS] token is inserted at the beginning of each sentence to utilize the [CLS] for classification and retrieval.

\begin{figure}[tbh!]
    \centering
    \includegraphics[width=0.8\linewidth]{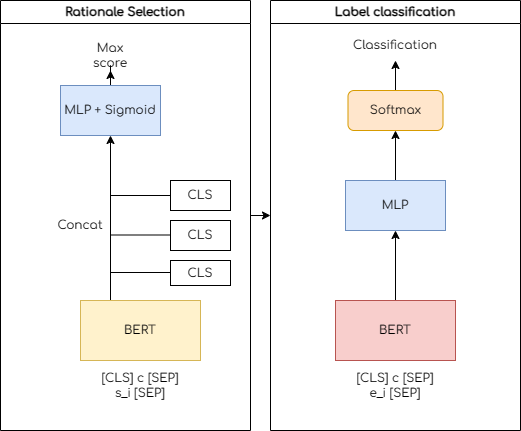}
    \caption{Pipeline for our approach}
    \label{fig:rightrecog}
\end{figure}

\subsection{Rationale Selection} \label{method:sent_select}
First, we will describe our rationale selection method. Given $s_1, s_2, ..., s_n$ as a sentence in the corpus, and $c$ as a claim, we define our rationale selection method in math notation as eq. \ref{eq:select-method-e}. 

\begin{equation} \label{eq:select-method-e}
    e = \max_{i \in \{1..n\}} P(s_i | c)
\end{equation}
where e is a top-1 sentence that retrieval from a corpus. 

So according to the eq. \ref{eq:select-method-e}, we should pick the top 1 sentence that has the same meaning as the claim.
This can be equivalent to a binary classification task when training, in which we can label 1 for evidence and 0 for otherwise for training purposes.
In the binary classification task, first we input [$c$ [SEP] $s_i$], in which the $c$ and $s_i$ is defined earlier.
Then it's passed into the BERT model and get the embedding of the [CLS] token, to form into $h_i = BERT($[$c$ [SEP] $s_i$]).
Moreover, not only do we take one 1 embedding [CLS] token in the last hidden state, but we also take 3 embedding of [CLS] tokens and then concatenate them to capture more information. 

After that, it's fed into the MLP Layer, which has 2 dense layers belonging to GELU\cite{DBLP:journals/corr/HendrycksG16} activation between them to calculate the probabilities of whether the sentence is or is not the evidence.
The output of MLP is then passed to the sigmoid function to ensure that the output is always between 0 and 1 as eq. \ref{MLP-layer}.

\begin{equation}
    p_i = \sigma(MLP(h_i))
    \label{MLP-layer}
\end{equation}
where $p_i$ is a probability of evidence.

During training time, binary cross entropy loss is used to calculate the loss between the probability and the ground truth label (1 for is evidence and 0 for otherwise) as eq. \ref{BCE-loss}.

\begin{equation}
    L = -\sum_{n=1}^{n} (y_i)*\log(\hat{y}) + (1- y_i)*\log(1 - \hat{y})
    \label{BCE-loss}
\end{equation}
where $\hat{y}$ is a ground truth label and $y$ is a probability of candidate sentence.

\subsection{Label classification} \label{subsec:label}
Next, our target is to predict a label with a retrieved sentence from the previous step. 
The normal approach would be given the candidate evidence $e$ and claim $c$, our goal is to find label $\hat{y}(c, a) \in \{SUPPORTED, REFUTED, NEI\}$. Given sentence s and claim c, we classify three label $\{SUPPORTED, REFUTED, NEI\}$ (we called that \textbf{Label Classification}) and assign it for the final result. 

With model classification, for each claim $c$ and evidence $v$, we form it as: $x = $ [$c$ [SEP] $e$]. 
Similar to the rationale selection tasks. 
Again, it's passed onto the BERT model to have a [CLS] token embedding $h_{[CLS]} = BERT(x)$. 
And then to the MLP layer with 2 dense layers and GELU\cite{DBLP:journals/corr/HendrycksG16} activation with softmax at the end to find the probability distribution of all labels as eq. \ref{softmax-res}.

\begin{equation}
    \hat{y} = softmax(MLP(h_{[CLS]}))
    \label{softmax-res}
\end{equation}
where : $y \in \{SUPPORTED, REFUTED, NEI\}$ 

The final result is the highest probability score from $\hat{y}$ as the final verdict $v = argmax(\hat{y})$.
During training, cross entropy loss is chosen, expressed as this equation \ref{CE-loss}.

\begin{equation}
    L = -\sum_{n=1}^{n} y_i\log(p_i)
    \label{CE-loss}
\end{equation}

\subsection{2-phase clasification}
\begin{figure}[tbh!]
    \centering
    \includegraphics[width=\linewidth]{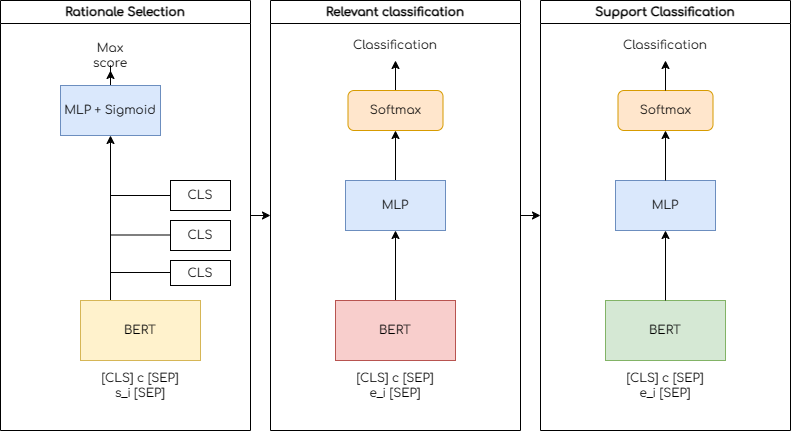}
    \caption{Pipeline for 3 phase}
    \label{fig:3phase}
\end{figure}

We also comparing this model to 2 phase model classification as described below. 
Instead of classifying three labels $\{SUPPORTED, REFUTED, NEI\}$ at the same time, it's split into two same classification models on \ref{subsec:label} with binary classification, one will be classified as $\{RELEVANT, N-REVEVANT\}$.
If the label is RELEVANT, one more classification task is needed to distinguish between SUPPORTED or REFUTED, otherwise, the label will be NEI.
This approach is given as figure \ref{fig:3phase}
Additionally, we find that if we use a dataset generated from \ref{method:sent_select} rather than at random, the performance will increase due to the learning of some minor cases about REFUTED and NEI.
The detailed result for this will be discussed in section 5.

\section{Experimental Results}\label{sec:experimental-results}
\subsection{Experiment setup}
The initial experiment was conducted on the ISE-DSC01 dataset in section \ref{sec:dataset}.
On the rationale selection module, we use PhoBERT\cite{nguyen-tuan-nguyen-2020-phobert} as the backbone and finetune it on the dataset. 
We utilize the model XLM-R \cite{conneau-etal-2020-unsupervised} with checkpoint \textit{xlm-roberta-large-xnli}\footnote{https://huggingface.co/joeddav/xlm-roberta-large-xnli}, which has been already finetune on XNLI\cite{conneau2018xnli} dataset, as the backbone for label classification.

From our experiment, we choose our best hyperparameter from finetuning. 
On each model, we show our detailed specifications of hyperparameters each module is provided in Table \ref{table:param1} for the rationale selection module and the label classification module, respectively. As GPU, we use single RTX 4090 24GB to train our model. 

\begin{table}[!tb]
\centering
\centering
\caption{rationale selection and Label classification modules configuration} \label{table:param1}
\begin{tabularx}{0.7\linewidth}{lXX}
\toprule
Hyperparameter\hspace{0.3cm}       & SS & LC          \\
\midrule
Learning rate & $5e^{-6}$, $1e^{-5}$, $2e^{-5}$     & $5e^{-6}$, $1e^{-5}$       \\
Batch size      & 8,16     & 4,8    \\
Epoch           & 2-4      & 3-4 \\
\bottomrule
\end{tabularx}
\end{table}

In this dataset, we will evaluate our result based on metric that belongs to the dataset. The main metric we use in this dataset is \textbf{Strict Match}, which is given as eq. \ref{metric-1}. 

\begin{equation}
    StrAcc = F(v, v') * F(e, e')
    \label{metric-1}
\end{equation}
where :
\begin{itemize}
\item Strict Accuracy denoted as StrAcc. 
\item $F(x,y) = 1$ if $x = y$ otherwise $\sigma(x,y) = 0$
\item $v, v'$ are predicted verdict and truth verdict
\item $e, e'$ predicted evidence and truth evidence
\end{itemize}
We also use \textbf{Accuracy} metric for both verdicts (denote as \textbf{\textit{Acc}}) and evidence (denote as \textbf{\textit{Acc@1}}), define as eq. \ref{metric-2}.
\begin{equation}
    Acc = \dfrac{N_{truth}}{N_{total}}
    \label{metric-2}
\end{equation}

\textbf{BERT\_FEVER}\cite{DBLP:journals/corr/abs-1910-02655}: uses 2 BERT separate modules: evidence extract and label classification. Additionally, they use TF-IDF for document retrieval but we won't use it on this dataset. Also, we will change the original BERT to the pre-trained mBERT because of the Vietnamese language support. 


\subsection{Results}

In this section, we present the result of our model, focusing on its accuracy on different test sets. 
Table \ref{result:table1} shows the performance of our models with one phase, with two phases, and the baseline model (BERT\_FEVER), 
compared on the public test and private test. 
According to table \ref{result:table1}, our model's performance on the private dataset is higher than the baseline model in all three metrics, significantly 28.33\% higher than BERT\_FEVER on Strict Acc.  
However, the result of 2-phase label classification is not as good as 1-phase on all three metrics. 
It shows that the 1-phase approach has higher capabilities in both determining the verdict (82.3\% on Acc, as compared to 77.39\% of the 2-phase one) and the evidence (76.82\% on Acc, as compared to 73.04\% of the 2-phase one).
Therefore, the outcome of the 1-phase approach is generally better than that of the 2-phase approach (75.11\% compared to 72\%).
On the public test, due to author had locked the public test submission, the 1-phase approach has yet to be tested.
However, with the result of the private test, it is confidently to say that the outcome of 1-phase approach 
is expected to be better than 2-phase approach.

\begin{table}[!tb]
    \centering
    \caption{Evaluation result (\%)} \label{result:table1}
    \vspace{0.1cm} 
    \begin{tabularx}{\linewidth}{@{}l *{6}{>{\centering\arraybackslash}X}@{}}
        \toprule
        & \multicolumn{3}{c}{Private Test} & \multicolumn{3}{c}{Public Test} \\
        \cmidrule(r){2-4} \cmidrule(l){5-7}
        & StrAcc & Acc & Acc@1 & StrAcc & Acc & Acc@1 \\
        \midrule
        BERT-FEVER & 46.28 & 51.50 & 63.03 & 69.67 & 75.05 & 70.23 \\
        Ours w/ 2 phase & 72.00 & 77.39 & 73.04 & 84.23 & 88.72 & 85.13 \\
        \textbf{Ours w/ 1 phase} & \textbf{75.11} & \textbf{82.30}  & \textbf{76.82} & - & - & - \\
        \bottomrule
    \end{tabularx}
\end{table}

Different backbone models have different capabilities on the Vietnamese language. To determine which one is well-suited with our task, we have experimented diffrent backbone models for the evidence retrieval task and the verdict classification task.
In table \ref{table:classifi}, we have experimented with three different backbone models for the claim verification module of our 1-phase classification model. 
According to the experimental result, we can see that with different claim verification modules, the outcome accuracy of our model exhibits substantial variability. 
The model xlm-robert-large-xnli performs better than the two other models on the evidence retrieval task (82.30\% for xlm-robert-large-xnli compared to 79.41\% for XLM-RoBERTa-Large and 74.22\% for PhoBERT-Large).
Because the XNLI dataset is a multilingual dataset for the Natural language inference task (classify 3 labels like our task, it has also been trained on a Vietnamese dataset), so fine-tuning action on it is expected to enhance our model's performance.

\begin{table}[!tb]
\centering
\begin{minipage}{.49\linewidth}
\centering
\caption{rationale selection accuracy (\%)} \label{table:sent}
\begin{tabular}{lcc}
\toprule
 Model & Acc@1 \\ \hline
\textbf{PhoBERT} & \textbf{59.43} \\ \hline
XLM-RoBERTa & 59.38 \\ \hline 
mBERT & 44.50 \\ \hline 
\end{tabular}
\end{minipage}\hfill
\begin{minipage}{.49\linewidth}
\centering
\caption{Classification accuracy (\%)} \label{table:classifi}
\begin{tabular}{lcc}
\toprule
 Model & Acc \\ \hline
XLM-RoBERTa-Large & 79.41 \\ \hline 
PhoBERT-Large & 74.22 \\ \hline 
\textbf{xlm-roberta-large-xnli} & \textbf{82.30} \\ \hline 
\end{tabular}
\end{minipage}
\end{table}

About the rationale selection's backbone model choices, we have experimented using three different pre-trained BERT models mBERT\cite{DBLP:journals/corr/abs-1810-04805}, PhoBERT\cite{nguyen-tuan-nguyen-2020-phobert} and XLM-RoBERTa\cite{conneau-etal-2020-unsupervised}. 
After evaluation, from the table \ref{table:sent}, the PhoBERT model performed slightly better than the XLM-RoBERTa model (+0.05\%) and significantly better than the mBERT model (+14.93\%). 
The fact that PhoBERT have better performance than XLM-RoBERTa on rationale selection task can be explained by the better accuracy in NLU (Natural Language Understanding) in the Vietnamese Language of PhoBERT according to the result of \cite{DBLP:journals/corr/abs-1910-02655}.

As for the label classification task, our model performance can be improved.
As table \ref{table:ex-dataset1}, our model can recognize correct labels even if the evidence sentence has been paraphrased. 
Besides, our model has some disadvantages in some cases of the dataset. 
According to table \ref{table:ex-dataset},  the human label would be NEI instead of REFUTED of our model. 
This can be explained that the claim and evidence from the corpus having almost the same sentence and only have one word difference (He and Khang). Our model cannot recognize it and therefore, a incorrect label is returned. 
\vspace{-0.5cm}
\begin{table}
\caption{True result} \label{table:ex-dataset1}
\tiny
\noindent \fbox{%
    \parbox{\textwidth}{%
        \textbf{Claim:} "Phương pháp giảm cân cấp tốc bằng cách ăn kiêng, chỉ ăn rau xanh, uống nước hay chỉ ăn một số loại thực phẩm nhất định được nhiều người áp dụng."\\
        \textit{Khang is from Bac Ninh and is studying in grade 12}
        \\
        \textbf{Evidence retrieval:} "\textbf{Nhiều người áp dụng phương pháp giảm cân cấp tốc bằng cách ăn kiêng, chỉ ăn rau xanh, uống nước hay chỉ ăn một số loại thực phẩm nhất định.}"\\
        \textit{\textbf{He is from Bac Ninh and is studying in grade 12}}
        \\
        \textbf{Label:} SUPPORTED \\
        \textbf{Human label:} \textbf{SUPPORTED}
    }%
    
}
\vspace{0.1cm}
\end{table}
\vspace{-0.5cm}
\begin{table}
\caption{False result} \label{table:ex-dataset}
\tiny
\noindent \fbox{%
    \parbox{\textwidth}{%
        \textbf{Claim:} "Khang ở Bắc Ninh, đang học lớp 12"\\
        \textit{Khang is from Bac Ninh and is studying in grade 12}
        \\
        \textbf{Evidence retrieval:} "\textbf{Em ở Gia Lai, đang học lớp 12.}"\\
        \textit{\textbf{He is from Bac Ninh and is studying in grade 12}}
        \\
        \textbf{Label:} REFUTED \\
        \textbf{Human label:} \textbf{NEI}
    }%
}
\vspace{0.1cm}
\end{table}



\section{Conclusion}\label{sec:conclusion}

The necessity for accurate claim verification has witnessed exponential growth in tandem with the proliferation of digital misinformation.
While significant strides have been made in claim verification for languages such as English, Chinese, and Danish, the direct applicability of these approaches to the Vietnamese language remains uncertain due to linguistic and cultural disparities.
This paper introduces an approach to address the challenges of Fact Verification using the Vietnamese dataset, aiming to enhance the accuracy of claim verification and evidence retrieval for the Vietnamese Fact Verification Dataset. 
To address these challenges, we propose a network architecture that integrates both sentence selection and classification modules. 
This combined approach aims to enhance the overall performance of the system.
To serve as the backbone of our architecture, we utilize pre-trained multilingual language models, namely PhoBERT and XLM-RoBERTa. 
These models were carefully chosen due to their demonstrated effectiveness in addressing the specific challenges posed by the problem at hand.
The experimental results demonstrate a significant improvement in our approach across all three metrics when compared to the baseline, with a substantial margin.
\subsubsection{Acknowledgements} 
This research is funded by Ho Chi Minh City University of Technology (HCMUT) – VNU-HCM under grant number 
SVOISP-2023-KH\&KTMT-44.
We acknowledge Ho Chi Minh City University of Technology (HCMUT), VNU-HCM for supporting this study.

\bibliographystyle{splncs04}
\bibliography{refs}

\begin{thebibliography}{10}
\providecommand{\url}[1]{\texttt{#1}}
\providecommand{\urlprefix}{URL }
\providecommand{\doi}[1]{https://doi.org/#1}

\bibitem{DBLP:journals/corr/BowmanAPM15}
Bowman, S.R., Angeli, G., Potts, C., Manning, C.D.: A large annotated corpus for learning natural language inference. CoRR  \textbf{abs/1508.05326} (2015), \url{http://arxiv.org/abs/1508.05326}

\bibitem{DBLP:journals/corr/ChenFWB17}
Chen, D., Fisch, A., Weston, J., Bordes, A.: Reading wikipedia to answer open-domain questions. CoRR  \textbf{abs/1704.00051} (2017), \url{http://arxiv.org/abs/1704.00051}

\bibitem{conneau-etal-2020-unsupervised}
Conneau, A., Khandelwal, K., Goyal, N., Chaudhary, V., Wenzek, G., Guzm{\'a}n, F., Grave, E., Ott, M., Zettlemoyer, L., Stoyanov, V.: Unsupervised cross-lingual representation learning at scale. In: Jurafsky, D., Chai, J., Schluter, N., Tetreault, J. (eds.) Proceedings of the 58th Annual Meeting of the Association for Computational Linguistics. pp. 8440--8451. Association for Computational Linguistics, Online (Jul 2020). \doi{10.18653/v1/2020.acl-main.747}, \url{https://aclanthology.org/2020.acl-main.747}

\bibitem{conneau2018xnli}
Conneau, A., Rinott, R., Lample, G., Williams, A., Bowman, S.R., Schwenk, H., Stoyanov, V.: Xnli: Evaluating cross-lingual sentence representations. In: Proceedings of the 2018 Conference on Empirical Methods in Natural Language Processing. Association for Computational Linguistics (2018)

\bibitem{dehaven-scott-2023-bevers}
DeHaven, M., Scott, S.: {BEVERS}: A general, simple, and performant framework for automatic fact verification. In: Akhtar, M., Aly, R., Christodoulopoulos, C., Cocarascu, O., Guo, Z., Mittal, A., Schlichtkrull, M., Thorne, J., Vlachos, A. (eds.) Proceedings of the Sixth Fact Extraction and VERification Workshop (FEVER). pp. 58--65. Association for Computational Linguistics, Dubrovnik, Croatia (May 2023). \doi{10.18653/v1/2023.fever-1.6}, \url{https://aclanthology.org/2023.fever-1.6}

\bibitem{DBLP:journals/corr/abs-1810-04805}
Devlin, J., Chang, M., Lee, K., Toutanova, K.: {BERT:} pre-training of deep bidirectional transformers for language understanding. CoRR  \textbf{abs/1810.04805} (2018), \url{http://arxiv.org/abs/1810.04805}

\bibitem{10013889}
Duong, H.T., Ho, V.H., Do, P.: Vietnamese fact checking based on the knowledge graph and deep learning. In: 2022 RIVF International Conference on Computing and Communication Technologies (RIVF). pp. 530--535 (2022). \doi{10.1109/RIVF55975.2022.10013889}

\bibitem{DBLP:journals/corr/HendrycksG16}
Hendrycks, D., Gimpel, K.: Bridging nonlinearities and stochastic regularizers with gaussian error linear units. CoRR  \textbf{abs/1606.08415} (2016), \url{http://arxiv.org/abs/1606.08415}

\bibitem{DBLP:journals/corr/abs-2011-03088}
Jiang, Y., Bordia, S., Zhong, Z., Dognin, C., Singh, M., Bansal, M.: Hover: {A} dataset for many-hop fact extraction and claim verification. CoRR  \textbf{abs/2011.03088} (2020), \url{https://arxiv.org/abs/2011.03088}

\bibitem{DBLP:journals/corr/abs-2108-11357}
Krishna, A., Riedel, S., Vlachos, A.: Proofver: Natural logic theorem proving for fact verification. CoRR  \textbf{abs/2108.11357} (2021), \url{https://arxiv.org/abs/2108.11357}

\bibitem{DBLP:journals/corr/abs-1910-09796}
Liu, Z., Xiong, C., Sun, M.: Kernel graph attention network for fact verification. CoRR  \textbf{abs/1910.09796} (2019), \url{http://arxiv.org/abs/1910.09796}

\bibitem{nguyen-tuan-nguyen-2020-phobert}
Nguyen, D.Q., Tuan~Nguyen, A.: {P}ho{BERT}: Pre-trained language models for {V}ietnamese. In: Cohn, T., He, Y., Liu, Y. (eds.) Findings of the Association for Computational Linguistics: EMNLP 2020. pp. 1037--1042. Association for Computational Linguistics, Online (Nov 2020). \doi{10.18653/v1/2020.findings-emnlp.92}, \url{https://aclanthology.org/2020.findings-emnlp.92}

\bibitem{norregaard-derczynski-2021-danfever}
N{\o}rregaard, J., Derczynski, L.: {D}an{FEVER}: claim verification dataset for {D}anish. In: Dobnik, S., {\O}vrelid, L. (eds.) Proceedings of the 23rd Nordic Conference on Computational Linguistics (NoDaLiDa). pp. 422--428. Link{\"o}ping University Electronic Press, Sweden, Reykjavik, Iceland (Online) (May 31--2 Jun 2021), \url{https://aclanthology.org/2021.nodalida-main.47}

\bibitem{DBLP:journals/corr/abs-2010-11930}
Pradeep, R., Ma, X., Nogueira, R.F., Lin, J.: Scientific claim verification with {VERT5ERINI}. CoRR  \textbf{abs/2010.11930} (2020), \url{https://arxiv.org/abs/2010.11930}

\bibitem{DBLP:journals/corr/abs-2202-02646}
Rana, A., Khanna, D., Singh, M., Ghosal, T., Singh, H., Rana, P.S.: Rerrfact: Reduced evidence retrieval representations for scientific claim verification. CoRR  \textbf{abs/2202.02646} (2022), \url{https://arxiv.org/abs/2202.02646}

\bibitem{DBLP:journals/corr/RiedelASR17}
Riedel, B., Augenstein, I., Spithourakis, G.P., Riedel, S.: A simple but tough-to-beat baseline for the fake news challenge stance detection task. CoRR  \textbf{abs/1707.03264} (2017), \url{http://arxiv.org/abs/1707.03264}

\bibitem{DBLP:journals/corr/abs-1910-02655}
Soleimani, A., Monz, C., Worring, M.: {BERT} for evidence retrieval and claim verification. CoRR  \textbf{abs/1910.02655} (2019), \url{http://arxiv.org/abs/1910.02655}

\bibitem{DBLP:journals/corr/abs-1803-05355}
Thorne, J., Vlachos, A., Christodoulopoulos, C., Mittal, A.: {FEVER:} a large-scale dataset for fact extraction and verification. CoRR  \textbf{abs/1803.05355} (2018), \url{http://arxiv.org/abs/1803.05355}

\bibitem{DBLP:journals/corr/abs-2004-14974}
Wadden, D., Lo, K., Wang, L.L., Lin, S., van Zuylen, M., Cohan, A., Hajishirzi, H.: Fact or fiction: Verifying scientific claims. CoRR  \textbf{abs/2004.14974} (2020), \url{https://arxiv.org/abs/2004.14974}

\bibitem{DBLP:journals/corr/Wang17j}
Wang, W.Y.: "liar, liar pants on fire": {A} new benchmark dataset for fake news detection. CoRR  \textbf{abs/1705.00648} (2017), \url{http://arxiv.org/abs/1705.00648}

\bibitem{DBLP:journals/corr/WilliamsNB17}
Williams, A., Nangia, N., Bowman, S.R.: A broad-coverage challenge corpus for sentence understanding through inference. CoRR  \textbf{abs/1704.05426} (2017), \url{http://arxiv.org/abs/1704.05426}

\end{thebibliography}

\end{document}